%% file: templateArxiv.tex
\documentclass{article}

\usepackage{PRIMEarxiv}

\usepackage[utf8]{inputenc} 
\usepackage[T1]{fontenc}    
\usepackage{hyperref}       
\usepackage{url}            
\usepackage{booktabs}       
\usepackage{amsfonts}       
\usepackage{nicefrac}       
\usepackage{microtype}      
\usepackage{lipsum}
\usepackage{fancyhdr}       
\usepackage{graphicx}       
\graphicspath{{media/}}     

\usepackage{amsmath}
\usepackage{booktabs}
\usepackage{tabularx}
\usepackage{threeparttable}

\usepackage{tcolorbox}
\tcbuselibrary{breakable}
\usepackage{listings}
\usepackage{xcolor}

\definecolor{promptgray}{RGB}{245,245,245}
\definecolor{promptheader}{RGB}{210,210,210}
\definecolor{promptborder}{RGB}{160,160,160}

\newcommand{\datasetname}{\textsc{ThReadMed-QA}}

\title{\datasetname: A Multi-Turn Medical Dialogue Benchmark from Real Patient Questions}

\author{
  Monica Munnangi{*}\\
  Khoury College of Computer Sciences \\
  Northeastern University \\
  Boston \\
   \And
  Saiph Savage \\
  Khoury College of Computer Sciences \\
  Northeastern University \\
  Boston \\
}

\begin{document}
\maketitle

\begingroup\def\thefootnote{*}\footnotetext{CorrecpondingAuthor: \texttt{munnangi.m@northeastern.edu}}\endgroup

\input{latex/00_abstract}

\input{latex/01_introduction}
\input{latex/02_related_work}
\input{latex/03_dataset}

\input{latex/04_methods}
\input{latex/05_results}
\input{latex/06_discussion}

\input{latex/07_limitations}


\bibliographystyle{unsrt}  
\bibliography{references}  

\newpage
\appendix

\input{latex/08_appendix}

\end{document}

%% file: latex/00_abstract.tex
\begin{abstract}
Medical question-answering benchmarks predominantly evaluate single-turn exchanges, failing to capture the iterative, clarification-seeking nature of real patient consultations. We introduce \datasetname, a benchmark of 2,437 fully-answered patient--physician conversation threads extracted from r/AskDocs, comprising 8,204 question--answer pairs across up to 9 turns. Unlike prior work relying on simulated dialogues, adversarial prompts, or exam-style questions, \datasetname  \, captures authentic patient follow-up questions and verified physician responses, reflecting how patients naturally seek medical information online. We evaluate five state-of-the-art LLMs---GPT-5, GPT-4o, Claude Haiku, Gemini~2.5~Flash, and Llama~3.3~70B---on a stratified test split of 238 conversations (948 QA pairs) using a calibrated LLM-as-judge rubric grounded in physician ground truth. Even the strongest model, GPT-5, achieves only 41.2\% fully-correct responses. All five models degrade significantly from turn~0 to
turn~2 ($p < 0.001$), with wrong-answer rates roughly tripling by the third turn. We identify a fundamental tension between single-turn capability and
multi-turn reliability: models with the strongest initial performance (GPT-5: 75.2; Claude Haiku: 72.3 out of 100) exhibit the steepest declines by turn~2 (dropping 16.2 and 25.0 points respectively), while weaker models plateau or marginally improve. We also introduce metrics to quantify the (1) consistency of the models across turns: Conversational Consistency Scores (CCS) and (2) and the rate at which the error compound Error Propagation Rate (EPR). CCS reveal that nearly one in three Claude Haiku’s conversations swing between a fully correct and a completely wrong response within the same thread. EPR shows that a single wrong turn raises the probability of a subsequent wrong turn by 1.9--6.1$\times$ across all models. We release the code and dataset at \url{https://github.com/monicamunnangi/ThReadMed-QA}.
\end{abstract}

%% file: latex/01_introduction.tex
\section{Introduction}
\label{sec:intro}

Large language models (LLMs) are increasingly used as interfaces for answering patient health questions, offering unprecedented accessibility to medical information and personalized guidance. Their strong performance on medical knowledge benchmarks and licensing-style examinations has fueled optimism about their potential to support clinical decision-making and patient education \cite{53083, singhal2023towards}. However, these evaluations are largely conducted under idealized conditions: single-turn prompts, carefully phrased questions, and the implicit assumption that users articulate medically coherent and accurate queries. As a result, existing benchmarks capture only a narrow slice of what it means for an LLM to behave safely and effectively in real-world patient-facing settings.

In practice, real patient queries bear little resemblance to the structured, unambiguous questions of medical licensing examinations. Real patient queries are iterative, emotionally charged, and often shaped by uncertainty, incomplete information, or incorrect causal beliefs. Patients may seek reassurance, embed false assumptions about diagnoses or medications directly into their questions, or introduce critical clinical details only gradually over multiple turns. Prior research has shown that such communication patterns can contribute to unsafe outcomes if not properly addressed \cite{bean2025clinical}. At the same time, the use of LLMs for health information is rapidly increasing \cite{FinneyRutten2019OnlineHI}. Yet current evaluation frameworks rarely test whether models can recognize and respond appropriately to these behaviors as they unfold across a conversation.

Recent work has begun probing LLM robustness through adversarial prompting and multi-turn stress testing \cite{manczak2025shallow, laban2025llms}. While valuable, these approaches typically rely on synthetic or exam-style prompts designed to elicit failures. They do not ground their evaluation in how patients naturally seek medical information. Consequently, a critical question remains unanswered: \textbf{How do LLMs perform as patient conversations unfold across multiple turns of realistic, patient-authored dialogue?} In an effort to answer this question, we create a dataset of patient-authored medical questions, their follow-ups, and physician responses to test LLMs' accuracy throughout the conversation. Recent work \cite{Stanwyck2026.02.12.26346164, Agrawal2025TheEI} emphasizes the need for realistic multi-turn evaluation benchmarks. Overall, these findings indicate that relying exclusively on single-turn, exam-style multiple-choice evaluations—particularly those with artificial or adversarial tests may exaggerate the effects of prior-chat bias. They underscore the importance of evaluating LLM performance in multi-turn, clinically realistic scenarios instead of depending solely on board-style benchmarks to characterize safety-related risks.

\begin{figure}[t!]
    \centering
    \includegraphics[width=0.8\textwidth]{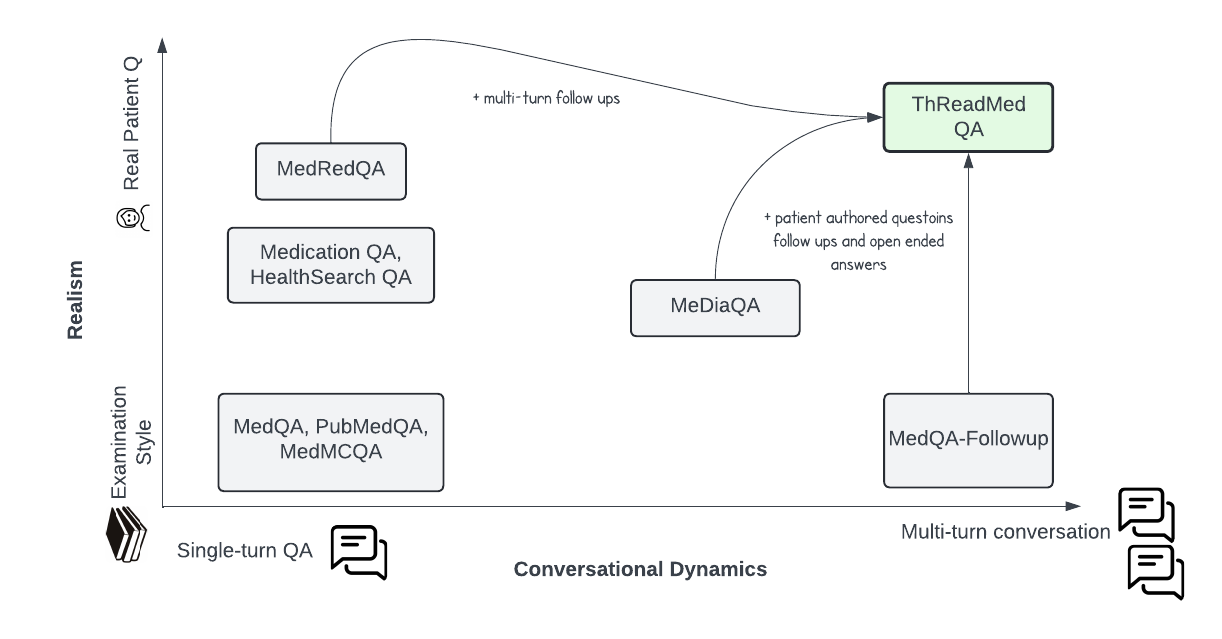}
    \caption{We compare our dataset with the existing medical QA datasets. MedQA \cite{jin2020diseasedoespatienthave}, PubMedQA \cite{jin2019pubmedqadatasetbiomedicalresearch}, and MedMCQA \cite{pal2022medmcqalargescalemultisubject} are derived from medical examinations and are multiple-choice questions. MedQA-Followup \cite{manczak2025shallow} and MeDiaQA \cite{suri2021mediaqaquestionansweringdataset} are adversarial follow-up questions with multiple choice options. MedRedQA \cite{nguyen2023medredqa}, MedicationQA \cite{abacha2019bridging} and HealthSearchQA \cite{53083} have patient-authored questions but in single-turn. Our dataset \datasetname is the first to have patient-authored questions and followups, along with physician responses.}
    \label{fig:dataset_placing}
\end{figure}

More concretely, in this work, we introduce \datasetname, a multi-turn evaluation framework designed to assess LLM robustness under real patient communication behaviors. Rather than treating patient questions as isolated inputs, \datasetname \ is constructed with patient-authored medical questions and their follow-ups. Figure~\ref{fig:dataset_placing} compares \datasetname with the existing medical QA datasets. We evaluate state-of-the-art LLMs and focus on multi-turn metrics to quantify their performance in real-world multi-turn contexts.

Evaluating generative medical responses remains an open challenge, particularly given the multiplicity of valid answers \cite{chang2024survey}. To address this, we incorporate scalable, rubric-based evaluation aligned with clinician judgment to ensure clinical relevance while maintaining reproducibility. By grounding multi-turn evaluation in real patient information-seeking trajectories rather than synthetic adversarial dialogues, \datasetname \, advances toward ecologically valid assessment of patient-facing LLMs.

By shifting the evaluation focus from ``Can the model answer correctly?'' to \textbf{``Does the model respond safely and appropriately as patient communication unfolds?''}, \datasetname \, reframes how clinical LLMs should be assessed. This shift is essential for understanding whether strong benchmark performance translates into reliable behavior in the real-world settings where these systems are increasingly deployed—and where the consequences of failure are most severe.
More concretely, here are our contributions:

\begin{enumerate}
    \item \textbf{A multi-turn patient evaluation dataset:} We construct a multi-turn benchmark grounded in patient-authored health questions and their follow-ups, reflecting how patients naturally seek medical information online. Our dataset also has physician authored responses for these questions.

    \item \textbf{Metrics to measure performance in a multi-turn setting}: We introduce metrics to capture distinct failure modes in multi-turn medical dialogue: (1) Turn-Level Degradation to measure how accuracy changes across conversation turns, Conversational Consistency Score (CCS) to quantify within-conversation reliability, and Error Propagation Rate (EPR) to measure whether a wrong response at one turn elevates the risk of failure at the next.

    \item \textbf{Evaluation of frontier LLMs under realistic patient interaction:} We evaluate state-of-the-art deployed LLMs using metrics tailored to multi-turn clinical settings, assessing their ability to maintain consistency, update reasoning, and safely address clinical concerns across turns, evaluated against the physicians response.

\end{enumerate}

%% file: latex/02_related_work.tex
\section{Related Work}

\paragraph{Evolution of medical benchmarking datasets}

Early benchmarks for medical question answering primarily emphasize factual recall and exam-style reasoning. Datasets such as MedQA \cite{jin2020diseasedoespatienthave}, MedMCQA \cite{pal2022medmcqalargescalemultisubject}, and PubMedQA \cite{jin2019pubmedqadatasetbiomedicalresearch} are constructed from medical licensing examinations or biomedical literature, enabling controlled evaluation of medical knowledge but offering limited insight into real-world patient interaction \cite{raji2025s, Agrawal2025TheEI}. Subsequent efforts sought to better reflect consumer-facing use cases by incorporating health-related search queries and community-authored questions, including Medication QA \cite{abacha2019bridging}, and HealthSearchQA \cite{53083}. There are several datasets constructed from online health forums, search logs, and tele-health consultations \cite{nguyen2023medredqa, abacha2019summarization}. One prior work stands out \cite{suri2021mediaqaquestionansweringdataset}, in which the authors collect multiple-choice questions (MCQs) from a Chinese telehealth portal, with follow-up questions and multiple-choice answers, does not reflect real-world complexities. These datasets improve coverage of lay language and consumer concerns, yet they continue to evaluate models largely through single-turn responses with static reference answers.

\paragraph{Multi-turn LLM Interaction in Medicine}

A growing literature highlights a disconnect between how medical LLMs are evaluated and how they are used in practice \cite{Agrawal2025TheEI, Stanwyck2026.02.12.26346164}. Systematic reviews show that the majority of medical LLM evaluations rely on synthetic or exam-style data, with minimal use of real clinical or patient-generated text \cite{Bedi2024.04.15.24305869}. These evaluations often assume complete and coherent input, overlooking the fragmented, uncertain, and iterative nature of real patient communication.

Recent work in the general-domain LLM literature demonstrates that model performance can degrade sharply when moving from single-turn to multi-turn interactions \cite{laban2025llms}. In medical settings, this brittleness is especially concerning: models that appear competent on static questions may fail to maintain consistency, challenge incorrect premises, or preserve appropriate safety guidance as a conversation unfolds. Closest to our work are studies that examine multi-turn or misconception-aware evaluation in specific clinical subdomains \cite{laban2025llms, manczak2025shallow}. However, these evaluations typically involve short, decontextualized prompts and do not model real patient information-seeking trajectories. Our work addresses this gap by explicitly evaluating longitudinal patient–LLM interactions grounded in authentic patient behavior (figure~\ref{fig:dataset_placing}).

\paragraph{Evaluation Methodology and LLM-as-a-Judge}

Evaluating generative medical responses remains challenging due to the diversity of acceptable answers and the difficulty of defining a single ground truth. Traditional overlap-based metrics and learned semantic scorers capture surface similarity but correlate weakly with expert judgments of safety and clinical usefulness \cite{chang2024survey}. Recent work increasingly combines multiple metric families with human evaluation and model-based judges to scale assessment \cite{zheng2023judgingllmasajudgemtbenchchatbot}. LLM-as-a-judge frameworks, when carefully designed and validated, have shown promise for capturing nuanced qualities such as factual correctness, reasoning quality, and alignment with domain-specific criteria \cite{pmlr-v298-munnangi25a}. Our evaluation framework follows this line of work, using structured rubrics to compare physician authored responses against LLM generated responses. 

%% file: latex/03_dataset.tex
\section{Dataset}

We introduce \datasetname \, a multi-turn medical question-answering benchmark derived from r/AskDocs, a Reddit community where patients seek medical advice from verified healthcare professionals. Unlike single-turn QA datasets, our benchmark captures extended patient–physician dialogues in which patients ask follow-up questions to clarify, refine, or challenge initial answers. This structure reflects real-world information-seeking behavior and enables evaluation of models on sustained reasoning, context retention, and handling of patient misconceptions.

The raw data is sourced from r/AskDocs \footnote{\url{https://www.reddit.com/r/AskDocs/}}. We restrict to posts from January 2015 through June 2023. The total number of posts was approximately 1.8 million. Posts are required to have at least one comment to ensure engagement. We apply standard preprocessing, following \cite{nguyen2023medredqa}; removal of moderator-removed or banned-user content, bot submissions, posts with fewer than five words, images-only posts, and deleted or removed content. URLs are stripped from text. The same cleaning pipeline is applied to comments, which are linked to posts via Reddit’s \texttt{link\_id} field.

\begin{figure}[h!]
    \centering
    \includegraphics[width=0.8\textwidth]{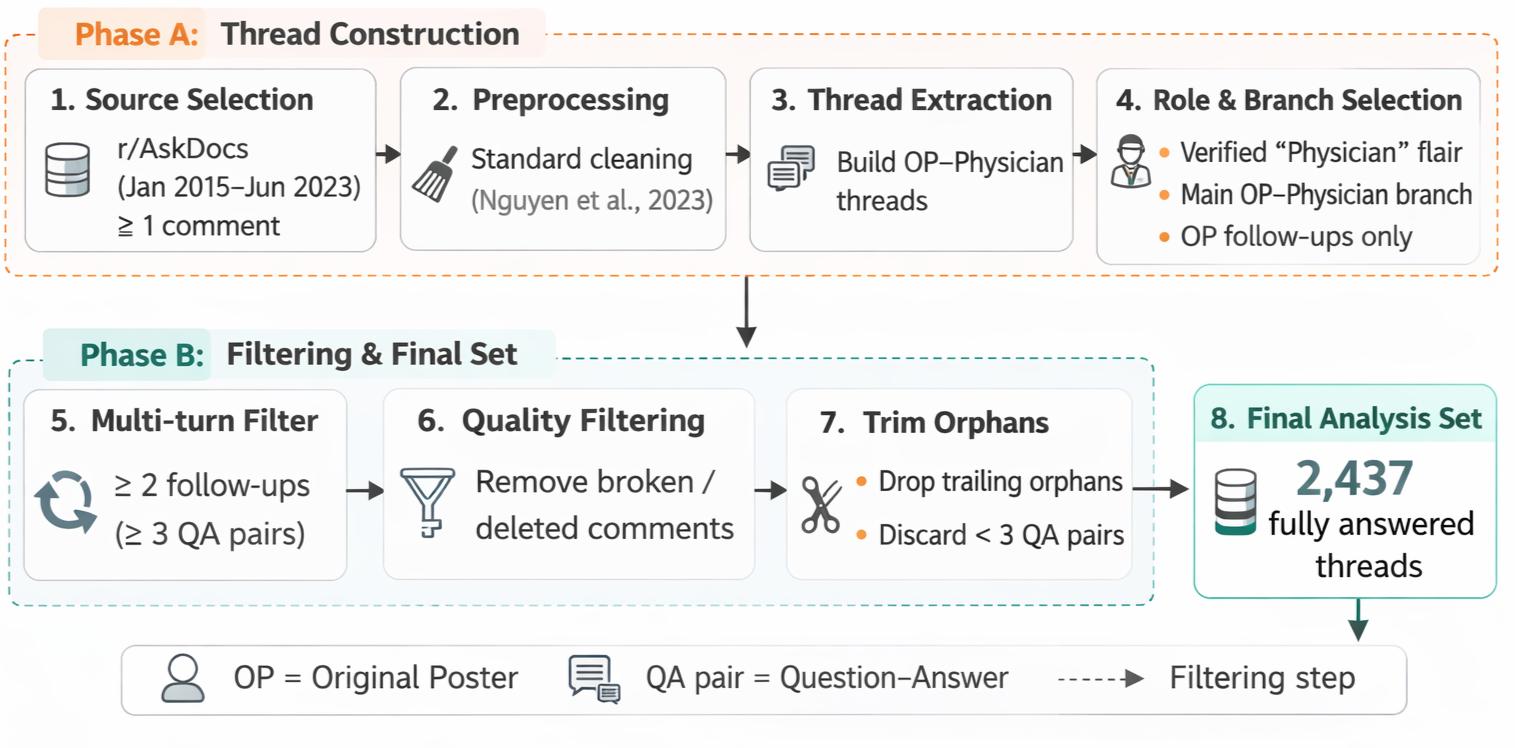}
    \caption{Dataset construction and filtering process.  We selected posts from r/AskDocs from Jan 2015 to Jun 2023, totaling approximately 1.8 million posts. After preprocessing and keeping the thread with the original poster (OP) and physician, we have the final set of 2,437 conversation threads.} 
    \label{fig:filtering}
\end{figure}

\subsection{Multi-turn Construction}

We extract linear patient–physician conversation threads using the following protocol:

\paragraph{Participant selection:} We retain only threads in which the original poster (OP) interacts exclusively with verified medical professionals. Answer authors must have a verified flair (e.g., Physician, Verified Physician) as designated by r/AskDocs moderators. Follow-up questions are restricted to the OP; we do not include follow-ups from other users or physicians.

\paragraph{Branch selection:} A single post may spawn multiple comment branches. We select the branch most relevant to the original question using a word-overlap heuristic (Jaccard-like overlap between the post text and the first physician's answer). This favors threads that stay on-topic and avoid tangents. We require at least two follow-up questions per thread to ensure genuine multi-turn structure. Threads with fewer follow-ups are discarded.

\paragraph{Quality filters:} Physician answers are filtered if they: (1) exhibit broken formatting (excessive [removed]/[deleted], garbled text); or (2) occur after the OP has deleted their account or removed their content mid-thread, in which case the entire branch is discarded.

\paragraph{Orphan Handling:} In the raw extraction, some questions receive no physician answer (e.g., the thread ends with an unanswered follow-up, or the answer was filtered). We refer to such question–answer pairs as \textit{orphans}. The initial dataset contains a substantial fraction of orphan pairs, with many threads ending in an unanswered final question. This dataset contains 9,741 conversation threads with 37,191 questions. Further, for a fully answered dataset, we select the conversation threads with no orphan questions. Threads with fewer than three QA pairs after trimming are dropped. This yields our primary benchmark: \textbf{2,437 conversation threads and 8,204 questions} with zero orphans, ensuring every conversation ends with a physician response.  We split this into an 80-10-10 train/val/test split, summarizing the process in figure~\ref{fig:filtering}. We report results on our test split.

\begin{table}[t]
\centering
\small
\begin{tabular}{p{4.5cm}rr}
\toprule
\textbf{Metric} & \textbf{Full Dataset} & \textbf{Fully Answered Dataset} \\
\midrule
\multicolumn{3}{l}{\textit{Scale}} \\
Total conversation threads & 9{,}741 & 2{,}437 \\
Total number of questions & 37{,}191 & 8{,}204 \\
Total number of answered questions & 21{,}595 (58.1\%) & 8{,}204 (100\%) \\
\addlinespace[3pt]
\multicolumn{3}{l}{\textit{Conversational depth}} \\
Follow-ups per thread (min / max) & 2 / 32 & 2 / 9 \\
Follow-ups per thread (mean / median) & 2.82 / 2.0 & 2.37 / 2.0 \\
Threads with $\geq$4 turns (\%) & 3{,}792 (38.9\%) & 622 (25.5\%) \\
Threads with $\geq$5 turns (\%) & 1{,}723 (17.7\%) & 185 (7.6\%) \\
\addlinespace[3pt]
\multicolumn{3}{l}{\textit{Length (num. tokens)}} \\
Question tokens (mean) & 128.6 & 141.7 \\
Answer tokens (mean) & 74.6 & 69.5 \\
Thread tokens (mean) & 656.6 & 711.2 \\
\bottomrule
\end{tabular}
\vspace{1em}
\caption{Comparison of the full multi-turn patient-physician QA dataset and the fully answered subset. Both datasets are derived from r/AskDocs threads with at least one verified medical professional respondent. The full dataset includes all threads with three or more QA turns (9,741 threads; 37,191 questions, 58.1\% answered). The fully answered subset retains only threads where every question has a physician response (2,437 threads; 8,204 questions). Token counts use tiktoken $ cl100k\_base $ (GPT-4/5). }
\label{tab:full_vs_subset}
\end{table}

\subsection{Dataset Characteristics}

Table~\ref{tab:full_vs_subset} presents summary statistics for our multi-turn medical QA benchmark. We report on two dataset variants: a full collection of 9,741 conversation threads (37,191 QA pairs) and a fully answered subset of 2,437 threads (8,204 QA pairs). The full dataset contains orphan questions (follow-ups that lack physician responses), reflecting realistic consultation patterns where physicians may not address every sub-question. The fully answered subset filters to threads in which \emph{every} question has a physician response, yielding a benchmark suitable for evaluation where ground-truth answers exist for all turns.

The data span 2015--2023, with peak activity in 2020--2022 coinciding with increased tele-health adoption during COVID-19. The majority of threads contain exactly three QA pairs (initial question plus two follow-ups): 61.1\% in the full dataset and 74.5\% in the fully answered subset. The median is 2 follow-ups per thread, with a long tail extending to 32 follow-ups in the full dataset (and up to 9 in the fully answered subset). In the full dataset of 37,191 QA pairs, 21,595 (58.1\%) have physician responses. The remaining 41.9\% (15,596 pairs) are mid-thread orphans---unanswered follow-ups that occur before later answered questions. In 7,241 threads, the final question has no answer. The fully answered subset of 2,437 threads contains 8,204 QA pairs with 100\% physician coverage.


%% file: latex/04_methods.tex
\section{Methods}
\label{sec:method}

\subsection{Experimental Pipeline}

We construct a stratified test split of \textbf{238 conversation threads} from the fully-answered subset of \datasetname, balanced by conversation length (short: $\leq$3 QA pairs; medium: 4-5; long: $\geq$6). Threads are sampled uniformly at random within each stratum, yielding \textbf{948 QA pairs} across 238 conversations for evaluations.

\subsection{Models}

We evaluate five state-of-the-art LLMs spanning the proprietary frontier and open-source categories:
GPT-5~\cite{OpenAI_2025} (snapshot: \texttt{gpt-5-2025-08-07}),
GPT-4o~\cite{openai2024gpt4ocard},
Claude Haiku 4.5~\cite{Claude-haiku} (snapshot: \texttt{claude-haiku-4-5-20251001}),
Gemini 2.5 Flash~\cite{comanici2025gemini25pushingfrontier},
and Llama~3.3--70B-Instruct~\cite{grattafiori2024llama}.
All models are evaluated zero-shot with no specific system prompt,  no few-shot examples or medical-domain specialization are applied, assessing out-of-the-box capability. Temperature is set to 0 for all models except GPT-5, which does not support deterministic decoding and is run at temperature 1. We only experiment with general-domain state-of-the-art models as numerous studies have shown domain-specific models often underperform \cite{jeong-etal-2024-medical, ceballos-arroyo-etal-2024-open} and are rarely deployed in patient-facing interfaces.

\subsection{Multi-Turn Inference}
\label{sec:method:inference}

For each thread, questions are presented sequentially with full conversation history, simulating a patient-facing medical assistant:

\begin{quote}
\small
\texttt{[User]:} $\langle$turn 0 patient question$\rangle$\\
\texttt{[Assistant]:} $\langle$model response$\rangle$\\
\texttt{[User]:} $\langle$turn 1 follow-up$\rangle$\\
\texttt{[Assistant]:} $\langle$model response$\rangle$\\
\quad $\vdots$
\end{quote}

Each model response conditions on all prior model outputs---the model's own
prior answers form part of the context for subsequent turns.

\subsection{Oracle-Physician Baseline}
\label{sec:method:oracle}

To isolate whether performance degradation stems from the models' conditioning on their own flawed prior responses (\emph{context poisoning}) or from
follow-up questions being intrinsically harder (\emph{question difficulty}), we run a controlled \emph{oracle-physician} condition. For turns 1+, all prior assistant turns are replaced with the verified
physician responses rather than the model's own outputs. Every other aspect of the evaluation is identical. This provides each model with the best possible conversational context and isolates how much degradation persists even with perfect prior turns. We evaluate 710 QA pairs (turns~1-5, 238 conversations) under this condition for all five models.

\subsection{LLM-as-a-Judge Evaluation}
\label{sec:method:judge}

We score model responses against the physician's ground truth using a calibrated
LLM-as-a-judge rubric, with GPT-4o as the judge. Each response is scored on a three-point scale (full prompt is provided in Appendix~\ref{apx:prompts}):

\begin{itemize}
    
    \item \textbf{1.0 (Correct):} Awarded only if \emph{all} of the following hold: the core medical explanation aligns with the physician's; recommended next steps match the physician's intent; any safety information or red flags the physician included are also present; no factually incorrect information is introduced; and the level of urgency matches the physician.
    
    \item \textbf{0.5 (Partially Correct):} The response is medically
    relevant but has exactly one of: a missing safety criterion or a red flag
    warning; a missing key diagnostic or management recommendation; excessive reassurance contradicting the physician's caution; or content that is correct but too generic to be actionable.
    Not awarded if factually incorrect information is present.

    \item \textbf{0.0 (Incorrect):} Awarded if any of the following hold:
    the response fails to engage with the actual question; addresses the wrong medical problem; contains a factual error that could harm the patient; is dangerously under-urgent relative to the physician's recommendation; or is dangerously over-urgent, leading to emergency care.
    
\end{itemize}

\subsection{Metrics}
\label{sec:method:metrics}

We report three metrics designed to characterize distinct failure modes
in multi-turn medical dialogue:

\paragraph{Turn-Level Degradation.}
Mean judge score and wrong-answer rate ($s = 0$) per turn group, with
bootstrap 95\% confidence intervals (10,000 resamples).
Individual turns 0, 1, 2 are reported separately ($n = 238$ each);
turns 3--5 are pooled into a \emph{late} group ($n = 210$) to maintain
adequate sample sizes; turns 6+ ($n = 24$) are reported as indicative.
Statistical significance of turn~0 vs.\ later groups is tested with the
one-sided Mann-Whitney U test.

\paragraph{Conversational Consistency Score (CCS).}

Turn-level averages measure whether a model performs well \emph{on average}, but not whether it is \emph{reliably} good within a single patient's consultation. A model that scores perfectly on turn~0 but completely wrong on turn~3 of the \emph{same} conversation is clinically problematic in a
way that mean scores obscure, a patient cannot know which of the model's responses to trust. CCS captures this within-conversation reliability by measuring how much a model's score varies across turns of the same thread.
A model with CCS $= 1$ gives the same quality response at every turn of every conversation; a model with CCS $= 0$ swings between perfect and completely wrong within every thread. Lower CCS indicates a model that is unpredictably inconsistent, regardless of how it performs on average.

We define CCS as:

\begin{equation}
  \mathrm{CCS} = 1 - \frac{1}{|\mathcal{C}|}\sum_{c \in \mathcal{C}}
  \bigl(\max_t s_{c,t} - \min_t s_{c,t}\bigr)
\end{equation}

where $s_{c,t}$ is the score at turn $t$ of conversation $c$ and
$\mathcal{C}$ is the set of conversations with $\geq 3$ turns
($n = 238$). CCS $= 1$ means a model never varies within a conversation;
CCS $= 0$ means it spans the full score range in every thread.

\paragraph{Error Propagation Rate (EPR).}
CCS measures the \emph{range} of a model's within-conversation variance but does not reveal whether failures are randomly scattered across turns
or whether a wrong response at one turn actively increases the risk of another wrong response at the next. EPR captures this compounding
behaviour. Intuitively, if a model misunderstands a patient's question at turn~2, does it recover at turn~3---or does the earlier failure
contaminate subsequent responses? A high EPR means errors are not isolated mistakes but the start of a sustained failure cascade, which
is particularly dangerous in a medical setting where a patient asking follow-up questions after a wrong answer is already in a vulnerable
position. Formally, we define EPR as the probability that a wrong response at turn $t$ is followed by a wrong response at turn $t+1$:

\begin{equation}
  EPR = P(s_{t+1} = 0 \mid s_t = 0)
\end{equation}

To assess whether this propagation is meaningfully elevated, we compare
EPR against the baseline wrong rate following a \emph{correct} turn,
$P(s_{t+1} = 0 \mid s_t = 1)$, and report the ratio of the two as the
\emph{amplification factor}. A large amplification factor indicates that errors compound substantially beyond what would be expected by chance, rather than reflecting a uniformly high wrong rate across all turn transitions.

%% file: latex/05_results.tex
\section{Results}
\label{sec:results}

We evaluate five LLMs on \textbf{948 QA pairs} across \textbf{238 multi-turn patient consultations}. Each response is scored $\{0, 0.5, 1.0\}$ by our LLM-as-judge rubric against verified physician answers. We report bootstrap 95\% CIs throughout; turns 3--5 are pooled (``late'') due to smaller sample sizes ($n = 40$--$109$ per turn), and turns 6+ ($n = 24$) are reported separately as indicative only.

\subsection{Overall Performance}
\label{sec:results:overall}

\begin{table}[h]
\centering
\setlength{\tabcolsep}{5pt}
\renewcommand{\arraystretch}{1.12}
\begin{tabular}{lccccc}
\toprule
\textbf{Model} & \textbf{Mean [95\% CI]} & \textbf{Correct (\%)} & \textbf{Wrong (\%)} & \textbf{Partially correct (\%)} & \textbf{Avg.\ Words} \\
\midrule
\emph{Physician} & \emph{---} & \emph{---} & \emph{---} & \emph{---} & \emph{62} \\
\midrule
GPT-5                & 65.0 [62.9, 67.0] & 41.2 & 11.3 & 47.5 & \emph{386} \\
Claude Haiku         & 54.4 [52.3, 56.6] & 28.1 & 19.2 & 52.7 & \emph{227} \\
GPT-4o               & 53.4 [51.7, 55.2] & 18.6 & 11.7 & 69.7 & \emph{264} \\
Gemini 2.5 Flash     & 45.7 [43.8, 47.6] & 12.9 & 21.5 & 65.6 & \emph{330} \\
Llama 3.3 70B        & 42.2 [40.7, 43.9] & 6.2 & 21.7 & 72.0 & \emph{323} \\
\bottomrule
\end{tabular}
\vspace{2em}
\caption{Overall performance (948 QA pairs, 238 conversations). Bootstrap 95\% CIs in brackets. Mean score is on a 0--100 scale (0 = wrong, 50 = partially correct, 100 = correct). All models produce responses 3.7--6.2$\times$ longer than those of physicians.}
\label{tab:overall}
\end{table}

We present the results in table~\ref{tab:overall}. We observe that GPT-5 leads at 65.0 (41.2\% correct), yet \textbf{no model answers the majority of questions correctly} against a physician benchmark.
The wrong-answer rate is non-monotonic in model capability: GPT-5 and GPT-4o both hold wrong rates near 11--12\%, while the three remaining models cluster at 19--22\%, indicating a reliability threshold between frontier and second-tier models not captured by the mean score. 
Response length negatively correlates with quality for GPT-4o ($r_s = -0.091$, $p = 0.004$) and is negligible for GPT-5 and Gemini ($|r_s| < 0.02$) suggesting that verbosity is not a quality signal.

\subsection{Turn-Level Degradation}
\label{sec:results:degradation}

\begin{table}[h]
\centering

\setlength{\tabcolsep}{4pt}
\renewcommand{\arraystretch}{1.12}
\begin{tabular}{lccccc}
\toprule
\textbf{Model}
  & \textbf{T0} {\small($n$=238)}
  & \textbf{T1} {\small($n$=238)}
  & \textbf{T2} {\small($n$=238)}
  & \textbf{T3--5} {\small($n$=210)}
  & \textbf{T6+\dag} {\small($n$=24)} \\
\midrule
GPT-5
  & 75.2 [71.4, 79.0]
  & 64.3 [60.1, 68.5]
  & 59.0 [54.6, 63.2]
  & 61.4 [56.9, 66.0]
  & 60.4 [47.9, 70.8] \\
Claude Haiku
  & 72.3 [68.5, 75.8]
  & 51.9 [47.7, 56.1]
  & 47.3 [43.1, 51.5]
  & 46.4 [41.9, 51.2]
  & 43.8 [31.2, 56.2] \\
GPT-4o
  & 58.2 [55.1, 61.3]
  & 51.8 [48.2, 55.3]
  & 50.2 [46.5, 53.9]
  & 52.6 [48.8, 56.3]
  & 50.0 [41.7, 58.3] \\
Gemini 2.5 Flash
  & 51.3 [48.1, 54.6]
  & 44.7 [41.2, 48.5]
  & 42.6 [38.9, 46.4]
  & 43.6 [39.3, 47.9]
  & 47.9 [37.5, 58.3] \\
Llama 3.3 70B
  & 47.1 [44.5, 49.6]
  & 40.5 [37.4, 43.7]
  & 40.1 [36.8, 43.5]
  & 41.0 [37.1, 44.8]
  & 43.8 [35.4, 52.1] \\
\bottomrule
\end{tabular}
\vspace{2em}
\caption{Mean judge score (0--100 scale) with bootstrap 95\% CIs by turn group. \dag: turns 6+ are indicative only ($n = 24$). See Appendix~\ref{apx:results}, table~\ref{tab:turns_pvalues} for full significance tests.}
\label{tab:turns}
\end{table}

\begin{figure}[t!]
    \centering
    \includegraphics[width=0.9\textwidth]{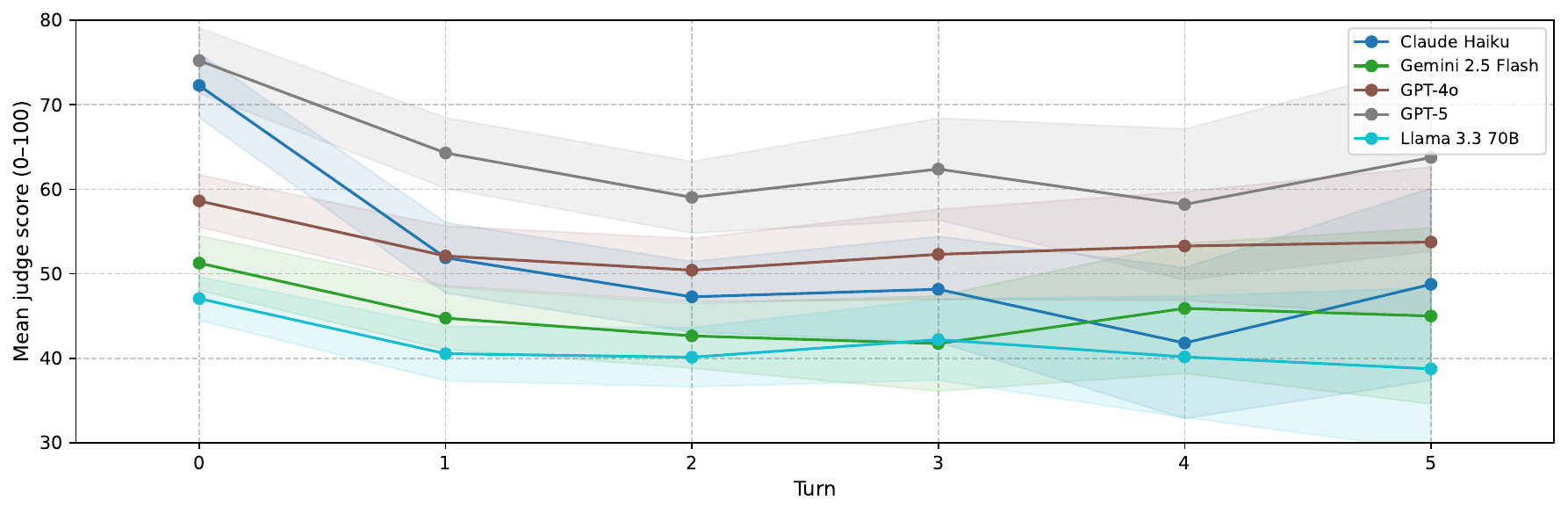}
    \caption{Mean judge score by conversation turn (turns 0-5, n = 238, 238, 238, 109, 61, 40 respectively). Shaded bands show 95\% confidence intervals. All models degrade from turn 0 to turn 2; the degradation is statistically significant across all models (p < 0.001, Mann-Whitney
U).} 
    \label{fig:turn_degradation}
\end{figure}

\textbf{Every model degrades significantly from turn~0 to all later turn groups} 
(Table~\ref{tab:turns}).
As observed in figure~\ref{fig:turn_degradation}, which plots mean score per turn, that all models have degrade in performance from turn~0 to turn~1. This degradation is independent of overall capability, as GPT-5, the strongest model, loses 16.2 points (75.2~$\to$~59.0) while GPT-4o loses only 8.0 points (58.2~$\to$~50.2).

Three qualitatively distinct profiles emerge.
\textbf{(1) Peak-and-cliff}: Claude Haiku starts highest (72.3) but suffers the steepest fall, 25.0 points by turn~2, a trajectory where strong initial performance provides false assurance of reliability.
\textbf{(2) Stable-floor}: GPT-4o and Gemini degrade sharply at turn~1 then plateau, suggesting a context floor below which they do not fall.
\textbf{(3) Monotonic decline}: Llama decreases at every turn with no recovery, reaching its minimum at turn~5 (38.8).
GPT-5 mirrors the stable-floor profile but shifted approximately 15--17 points higher at every turn; notably, the gap between GPT-5 and GPT-4o narrows from 17.0 points at turn~0 to 8.8 points by turn~2, indicating that multi-turn settings partially close the frontier gap.

\subsection{Conversational Consistency Score}
\label{sec:results:ccs}

\begin{table}[h]
\centering
\setlength{\tabcolsep}{5pt}
\renewcommand{\arraystretch}{1.12}
\begin{tabular}{lccccc}
\toprule
\textbf{Model} & \textbf{CCS [95\% CI]} & \textbf{Floor} & \textbf{Ceiling}
               & \textbf{Volatile (\%)} & \textbf{Degraded (\%)} \\
\midrule
GPT-5            & 51.5 [47.7, 55.3] & 39.1 & \textbf{87.6} & 16.8 & 57.1 \\
Claude Haiku     & 45.2 [41.0, 49.4] & 27.3 & 82.1          & \textbf{27.7} & \textbf{63.9} \\
GPT-4o           & 60.5 [56.7, 64.3] & 34.2 & 73.7          &  9.7 & 37.8 \\
Gemini 2.5 Flash & 57.1 [53.2, 61.1] & 25.0 & 67.9          & 13.0 & 39.9 \\
Llama 3.3 70B    & \textbf{63.9} [60.1, 67.4] & 22.9 & 59.0 &  8.0 & 38.7 \\
\bottomrule
\end{tabular}
\vspace{1em}
\caption{Conversational Consistency Score (0--100 scale) with bootstrap 95\% CIs (238 conversations, $\geq$3 turns). \textit{Floor}/\textit{Ceiling}: mean per-conversation worst- and best-case scores (0--100 scale).
\textit{Volatile}: conversation contains both a fully correct and a completely wrong response. \textit{Degraded}: later-turn mean $>$10 points below turn-0 score.}
\label{tab:ccs}
\end{table}

Table~\ref{tab:ccs} reveals a counter-intuitive ordering. Llama achieves the highest CCS (63.9) despite the lowest ceiling (59.0)---it is consistent precisely because it is \emph{consistently mediocre}. Claude Haiku has the lowest CCS (45.2) and the highest volatile rate (27.7\%): nearly one in three conversations swings between a fully correct and a completely wrong response within the same thread. GPT-5 has both the highest ceiling (87.6) and the highest floor (39.1), yet its CCS (51.5) is only third-highest because its within-conversation score range remains wide; a model that can reach 87.6 but also fall to 39.1 in the same conversation offers inconsistent clinical guidance. These patterns establish that \textbf{a high mean score or high ceiling does not guarantee consistent guidance across a consultation}: the model a patient experiences at turn~3 may be substantially worse than the one they encountered at turn~0, with no signal from the conversation itself about which to trust.

\subsection{Error Propagation Rate}
\label{sec:results:epr}

\begin{figure}[t!]
    \centering
    \includegraphics[width=0.8\textwidth]{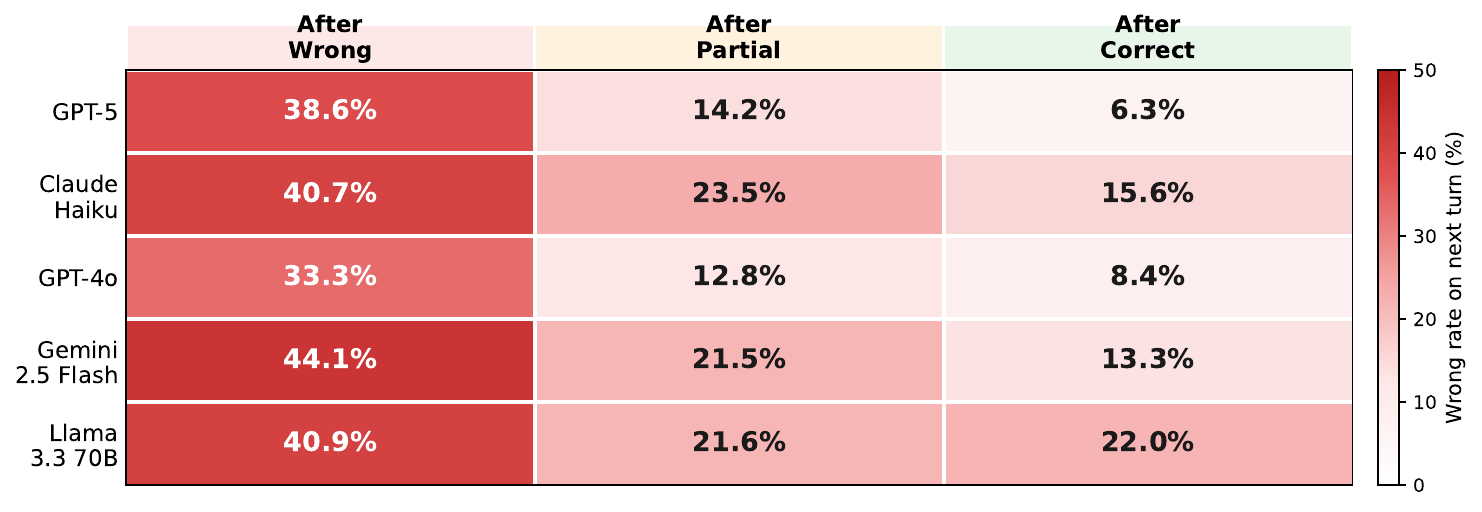}
    \caption{Cascade failure rates: percentage of wrong responses at turn t + 1 conditioned on the score at turn t. All models show severe cascade effects after a wrong turn, with failure rates up to 6.1× higher than after a correct turn.} 
    \label{fig:cascade}
\end{figure}

EPR is high across all models (35--44\%), and is 1.9--6.1$\times$ higher than the wrong rate following a correct turn (Table~\ref{tab:epr},
Figure~\ref{fig:cascade}).
Gemini has the highest EPR (44.1\%): a single off-track turn is most likely to derail its subsequent responses. Llama has the worst after-correct rate (22.0\%): even its sporadic correct answers are unstable and cannot be sustained. For GPT-5, EPR amplification is 6.1$\times$ meaning, once it goes wrong, it cascades as severely as any model despite having the lowest overall wrong rate.

\begin{table}[h]
\centering
\caption{Error Propagation Rate with bootstrap 95\% CIs.
\textit{After Correct}: wrong rate given prior correct turn.
\textit{Amplification}: EPR / after-correct wrong rate.}
\label{tab:epr}
\setlength{\tabcolsep}{6pt}
\renewcommand{\arraystretch}{1.12}
\begin{tabular}{lrrrr}
\toprule
\textbf{Model} & \textbf{EPR [95\% CI]} & \textbf{After Correct} & \textbf{Amplification} \\
\midrule
GPT-5            & 38.6\% [27.1, 50.0] &  6.3\% & 6.1$\times$ \\
Claude Haiku     & 40.7\% [31.9, 50.4] & 15.6\% & 2.6$\times$ \\
GPT-4o           & 35.1\% [24.7, 46.8] &  8.3\% & 4.2$\times$ \\
Gemini 2.5 Flash & \textbf{44.1\%} [36.0, 52.2] & 13.3\% & 3.3$\times$ \\
Llama 3.3 70B    & 40.9\% [32.6, 49.2] & \textbf{22.0\%} & 1.9$\times$ \\
\bottomrule
\end{tabular}
\end{table}

\subsection{Decomposing Degradation: Question Difficulty vs.\ Context Effects}
\label{sec:results:oracle}

Is degradation driven by models conditioning on their own flawed prior
responses (\emph{context poisoning}), or are follow-up questions intrinsically
harder? To isolate these effects, we run an \emph{oracle-physician} ablation:
all prior turns are replaced with verified physician responses, providing the
model with perfect conversational context. We evaluate 710 QA pairs
(turns~1--5, 238 conversations) under this condition.

\begin{table}[h]
\centering
\setlength{\tabcolsep}{5pt}
\renewcommand{\arraystretch}{1.12}
\begin{tabular}{lccccc}
\toprule
\textbf{Model} & \textbf{T0} & \textbf{Baseline T2} & \textbf{Oracle T2}
               & \textbf{Q-difficulty}$^\dagger$ & \textbf{Context effect} \\
\midrule
GPT-5            & 75.2 & 59.0 & 50.0 & 156\% of drop & $-9.0$\phantom{0} ($p < 0.001$) \\
Claude Haiku     & 72.3 & 47.3 & 50.8 &  86\% of drop & $+3.6$\phantom{0} ($p = 0.024$) \\
GPT-4o           & 58.2 & 50.2 & 45.4 & 160\% of drop & $-4.8$\phantom{0} ($p = 0.003$) \\
Gemini 2.5 Flash & 51.3 & 42.6 & 45.6 &  66\% of drop & $+2.9^\ddagger$\phantom{0} ($p = 0.36$) \\
Llama 3.3 70B    & 47.1 & 40.1 & 35.7 & 164\% of drop & $-4.4$\phantom{0} ($p = 0.003$) \\
\bottomrule
\end{tabular}
\vspace{1em}
\caption{Oracle-physician ablation at turn~2. \textit{Q-difficulty}: share of total degradation explained by question difficulty alone (baseline T0 $-$ oracle T2). \textit{Context effect}: oracle T2 $-$ baseline T2; negative values mean the oracle condition is \emph{worse} than baseline. $p$-values test baseline vs.\ oracle across turns~1--5 (Mann-Whitney U). {\footnotesize $^\dagger$ Values $>$100\% indicate the oracle condition scores \textit{below} baseline T2, implying the model's own prior responses
serve as net-positive scaffolding; question difficulty thus more than fully explains the total degradation. $^\ddagger$ Gemini 2.5 Flash oracle vs. baseline difference is not significant ($p = 0.36$); no directional claim is made for this model.} }
\label{tab:oracle}
\end{table}

The oracle results are counter-intuitive: \textbf{perfect prior context does not recover performance and in fact worsens it for three of five models} (GPT-5, GPT-4o, Llama; all $p \leq 0.003$, Table~\ref{tab:oracle}). Question difficulty accounts for the dominant share of total degradation
across most models (86--164\%): follow-up questions are intrinsically harder even with a perfect context, confirming that degradation cannot be attributed
primarily to cascading context errors. For GPT-5, GPT-4o, and Llama, the model's own prior responses appear to serve as helpful scaffolding---removing them hurts.
Claude Haiku is the exception: its oracle score improves modestly ($+$3.6, $p = 0.024$), suggesting partial context poisoning consistent with its peak-and-cliff degradation profile. \textbf{The primary driver of multi-turn degradation in medical dialogue is question difficulty, not context processing failure}---a finding that challenges purely architectural explanations of multi-turn unreliability.

\paragraph{Coverage.}
An oracle ensemble of all five models still fails on \textbf{47.4\% of QA pairs} (449/948): no model answers correctly on these turns.
GPT-5 uniquely solves 132 pairs (13.9\%) that no other model handles correctly; Claude Haiku adds 56 (5.9\%). The collective failure reveals a structural gap between current LLMs and physician-level multi-turn consultation that is not closed by model diversity.

%% file: latex/06_discussion.tex
\section{Discussion}
\label{sec:discussion}

\paragraph{The evaluation gap in medical AI.}
Benchmark performance on medical licensing examinations has become a de facto proxy for clinical readiness, yet it measures a capability that bears little resemblance to how patients actually seek care. Real consultations are iterative, emotionally inflected, and marked by incomplete disclosure: patients reveal critical details across multiple exchanges, embed false beliefs in their questions, and use social turns---``That makes sense,'' ``Thank you, but''---not to close the conversation
but to signal readiness for the next piece of guidance. Single-turn benchmarks, by construction, evaluate none of this. They reward encyclopedic recall and penalize nothing that emerges only over time: context drift, cascading failures, or the failure to proactively volunteer safety information when a patient moves on. The result is a systematic overestimate of clinical reliability. A model that scores 90\% on MedQA \cite{OpenAI_2025} and 41\% on the first turn of a real patient consultation, as GPT-5 does here, is not a model with a minor robustness gap; it is a model whose benchmark performance is largely uninformative about its behavior in the settings where it is increasingly deployed.

\datasetname is the first large-scale benchmark of authentic, multi-turn patient--physician dialogue evaluated against verified physician responses as ground truth. By extracting 2,437 fully answered conversation threads from r/AskDocs; where every follow-up has a physician response, and every patient turn was authored under the communicative pressures of real
information-seeking, we ground evaluation in ecological validity that synthetic and adversarial benchmarks cannot replicate. The rubric-based LLM-as-a-Judge framework, calibrated against explicit criteria for safety, urgency, and proactive information volunteering, makes scores interpretable beyond raw accuracy: a score of 0.5 is not a near-miss but a clinically meaningful failure, and a score of 0.0 on a social turn that a physician used to deliver safety criteria is a patient safety event.

The five models we evaluate span the full current capability range, from GPT-5 to Llama~3.3~70B, yet all share a structural failure: \emph{performance degrades universally and significantly as conversations progress.} The wrong-answer rate roughly triples from turn 0 to turn 2 across all models ($p < 0.001$ for each), meaning the population of patients most in need of reliable guidance, those asking follow-ups because the initial answer was unclear or alarming, is also the population most likely to receive a wrong response. That 47.4\% of QA pairs receive no fully correct response from any of the five models, including GPT-5, reveals that the bottleneck is not solved by scale alone.

Two findings are particularly instructive. First, the \emph{single-turn - multi-turn robustness trade-off}: the models with the strongest single-turn performance (GPT-5, Claude Haiku) exhibit the steepest multi-turn degradation, losing 11\% and 27.5\% of their turn-0 performance by turn 2, respectively. This reflects a structural tension between the optimization objectives that produce strong single-turn responses with comprehensiveness, confidence, and verbose elaboration but lack the contextual discipline required to track an evolving patient narrative across turns. A model trained to produce maximally informative single-turn answers will tend to anchor on its own prior response, overfill the context with assumptions, and misread social turns as opportunities for reassurance rather than as clinical handoffs. GPT-4o loses only 8.0 points from turn~0 to turn~2 despite a weaker starting position, suggesting that a lower-verbosity, more conservative response style actually confers a multi-turn advantage, a finding with direct implications for how medical AI systems should be prompted and fine-tuned.

Second, the \emph{cascade failure} pattern: a single wrong turn elevates the probability of a wrong next turn by 1.9--6.1$\times$ across all models. For patients with complex or anxiety-driven presentations, precisely the population that generates multi-turn threads, this means that early misreads compound into sustained unreliability.

\paragraph{Implications for the research community.}
\datasetname~is designed to support several research directions beyond the baselines reported here. The dataset enables direct study of \emph{context
retention}: does a model appropriately update its clinical reasoning when a patient reveals, at turn~3, that they are pregnant? Does it track prior symptom descriptions without re-asking for them? These are questions about memory and coherence that the benchmark's multi-turn structure makes
measurable. Beyond dataset use, the three metrics we introduce: Turn-Level Degradation, Conversational Consistency Score (CCS), and Error Propagation
Rate (EPR), provide a reusable evaluation vocabulary for multi-turn medical dialogue. CCS and EPR in particular surface failure modes that aggregate
accuracy scores obscure: a model can perform well on average while being dangerously inconsistent within individual consultations, or while allowing
single failures to cascade. We encourage future work to report these metrics alongside standard accuracy measures when evaluating patient-facing LLMs,
and to use them as optimization targets for fine-tuning models that must sustain reliability across extended conversations. Our LLM-as-judge rubric,
with its explicit criteria for urgency calibration and safety-information volunteering, offers a complementary evaluation scaffold that can be applied
to new models, new dialogue settings, or fine-tuned variants without human re-annotation at each iteration.

%% file: latex/07_limitations.tex
\section{Limitations and Future Work}
The benchmark's ground truth is physician responses from an online forum, which are not equivalent to formal clinical documentation and may themselves vary in completeness. Future work should validate judge scores against structured physician ratings on a held-out sample, and extend evaluation to non-English conversations and multimodal inputs (e.g., patient-uploaded images) that are common in real consultations. The evaluation also does not yet capture \emph{longitudinal} patient behavior, the same patient returning days later with a new symptom. Nor does it capture the specific failure mode of models that correctly answer but in a register so clinical that patients cannot act on the guidance. Addressing these dimensions will require both richer annotation and evaluation rubrics that incorporate patient comprehension, not only physician alignment.

%% file: latex/08_appendix.tex
\section{Additional Results}
\label{apx:results}

\subsection{Turn wise degradation}
We show turn-level degradation fro turn 0 to later turns with significance tests. 

\begin{table}[h]
\centering
\small
\setlength{\tabcolsep}{4pt}
\renewcommand{\arraystretch}{1.15}
\begin{tabular}{lccccc}
\toprule
\textbf{Model}
  & \textbf{T0} {\small($n$=238)}
  & \textbf{T1} {\small($n$=238)}
  & \textbf{T2} {\small($n$=238)}
  & \textbf{T3--5} {\small($n$=210)}
  & \textbf{T6+\dag} {\small($n$=24)} \\
\midrule
GPT-5
  & 75.2 [71.4, 79.0]
  & 64.3 [60.1, 68.5]$^{***}$
  & 59.0 [54.8, 63.2]$^{***}$
  & 61.4 [56.9, 66.0]$^{***}$
  & 60.4 [50.0, 72.9]$^{**}$ \\
Claude Haiku
  & 72.3 [68.5, 75.8]
  & 51.9 [47.7, 56.1]$^{***}$
  & 47.3 [43.3, 51.5]$^{***}$
  & 46.4 [41.9, 51.2]$^{***}$
  & 43.8 [31.2, 56.2]$^{***}$ \\
GPT-4o
  & 58.2 [55.1, 61.3]
  & 51.8 [48.4, 55.3]$^{**}$
  & 50.2 [46.5, 53.9]$^{***}$
  & 52.6 [48.8, 56.3]$^{*}$\phantom{**}
  & 50.0 [41.7, 58.3]\phantom{$^{**}$} ns \\
Gemini 2.5 Flash
  & 51.3 [48.1, 54.6]
  & 44.7 [41.0, 48.3]$^{**}$
  & 42.6 [38.9, 46.4]$^{***}$
  & 43.6 [39.5, 47.6]$^{**}$
  & 47.9 [37.5, 58.3]\phantom{$^{**}$} ns \\
Llama 3.3 70B
  & 47.1 [44.5, 49.6]
  & 40.5 [37.4, 43.7]$^{***}$
  & 40.1 [36.8, 43.5]$^{***}$
  & 41.0 [37.1, 44.5]$^{**}$
  & 43.8 [35.4, 52.1]\phantom{$^{**}$} ns \\
\bottomrule
\end{tabular}
\vspace{1em}
\caption{Mean judge score (0--100 scale) with bootstrap 95\% CIs by turn group,
with statistical significance of turn~0 vs.\ each later group
(Mann-Whitney U, one-sided: turn~0 $>$ later group).
\dag: turns 6+ are indicative only ($n = 24$).
$^{*}p<0.05$, $^{**}p<0.01$, $^{***}p<0.001$, ns: not significant.}
\label{tab:turns_pvalues}
\end{table}

\section{Inference and Evaluation Costs}

We report API costs for generating responses for 244 conversation threads and LLM-as-a-Judge evaluation in tables~\ref{tab:benchmark-inference} and ~\ref{tab:benchmark-judge}.

\begin{table}[ht]
\centering
\begin{tabular}{lrrr}
\toprule
  Model & Input (\$) & Output (\$) & Total (\$) \\
\midrule
  claude-haiku & 0.45 & 2.43 & 2.88 \\
  gemini-2.5-flash & 0.14 & 1.21 & 1.35 \\
  llama33\_70b & 0.40 & 0.43 & 0.83 \\
  gpt-4o & 2.26 & 7.28 & 9.55 \\
  gpt-5 & 0.57 & 4.85 & 5.42 \\
\bottomrule
\end{tabular}
\vspace{1em}
\caption{Inference cost for 5 models on 238 threads (948 QA pairs).}
\label{tab:benchmark-inference}
\end{table}

\begin{table}[ht]
\centering

\begin{tabular}{lr}
\toprule
  Judge model & Cost (\$) \\
\midrule
  GPT-4o & 30.98 \\
\bottomrule
\end{tabular}
\vspace{1em}
\caption{LLM-as-judge cost (GPT-4o) for all 5 models' outputs (4740 judgments).}
\label{tab:benchmark-judge}
\end{table}

\section{Prompts}
\label{apx:prompts}

\begin{tcolorbox}[
    colback=promptgray,
    colframe=promptborder,
    title=\textbf{Evaluation Prompt},
    breakable,
    boxrule=0.8pt,
    arc=4pt
]

\section*{LLM-as-a-Judge Evaluation Prompt}

To evaluate model responses against physician ground truth answers, we use a rubric-based grading framework. The full system and user prompts are provided below for reproducibility.

\subsubsection*{System Prompt}

You are an expert medical evaluator assessing AI-generated responses to patient medical questions.

Your task is to compare an AI's answer to a verified physician's answer using a 3-point scale. Your evaluation must focus on patient outcomes: would this response lead the patient to the same understanding and actions as the physician's response? \\

\textbf{Scoring Rubric — read carefully before scoring:} \\

\textbf{Score 1.0 — Correct} \\

Award 1.0 ONLY if ALL of the following are true:
\begin{itemize}
    \item The core medical explanation or diagnosis aligns with the physician's
    \item Recommended next steps / management match the physician's intent
    \item Any safety information or red flags the physician included are also present
    \item The response does not introduce factually incorrect medical information
    \item The level of urgency matches the physician (neither over- nor under-alarming)
\end{itemize} 

Note: The AI does not need to use identical wording or cover every detail. A concise response can score 1.0 if it covers the essential points. \\

\textbf{Score 0.5 — Partially Correct} \\

Award 0.5 when the response is medically relevant and partially useful, but has ONE of these problems:
\begin{itemize}
    \item Missing a specific safety criterion or red-flag warning the physician included
    \item Missing a key diagnostic or management recommendation that changes patient action
    \item Over-reassuring the patient in a way that contradicts the physician's level of caution
    \item The core content is correct but overly generic and provides little actionable guidance
\end{itemize}

Do NOT award 0.5 if the response introduces factually incorrect medical information (score 0.0 instead) or if the omission is a minor detail that would not change patient action. \\

\textbf{Score 0.0 — Incorrect} \\
 
Award 0.0 if ANY of the following are true:
\begin{itemize}
    \item The response fails to engage with the actual medical question
    \item The response addresses the wrong medical problem or body system
    \item The response contains a factual medical error that could harm the patient
    \item The response is dangerously under-urgent relative to the physician's recommendation
    \item The response is dangerously over-urgent in a way that may cause unnecessary emergency care
\end{itemize} 
\vspace{1em}

\textbf{IMPORTANT NOTES:}
\begin{itemize}
    \item Verbosity is not quality.
    \item Generic deferral ("consult your doctor") does not satisfy a safety requirement.
    \item If the physician proactively provides important information not directly requested, the AI is expected to do the same.
\end{itemize}

\vspace{1em}

\subsubsection*{User Prompt}
\vspace{0.5em}

\textbf{Patient Question:} \{question\}

\textbf{Physician Answer (Ground Truth):} \{physician\_answer\}

\textbf{AI Answer:} \{model\_answer\}

\vspace{0.5em}

Before scoring, briefly identify:
\begin{enumerate}
    \item What is the core medical point the physician is making?
    \item Does the AI response address this core point?
    \item Are there safety/urgency elements in the physician's answer? Does the AI include them?
\end{enumerate}

Then provide your score. \\

\textbf{Output as JSON:}

\begin{verbatim}
{"SCORE": <0.0, 0.5, or 1.0>, 
 "REASON": "<one sentence explaining the specific gap or alignment>"}
\end{verbatim}

\end{tcolorbox}